# Direct and Indirect Effects

Judea Pearl
Cognitive Systems Laboratory
Computer Science Department
University of California, Los Angeles, CA 90024
*judea@cs.ucla.edu*

## Abstract

The direct effect of one event on another can be defined and measured by holding constant all intermediate variables between the two. Indirect effects present conceptual and practical difficulties (in nonlinear models), because they cannot be isolated by holding certain variables constant. This paper presents a new way of defining the effect transmitted through a restricted set of paths, without controlling variables on the remaining paths. This permits the assessment of a more natural type of direct and indirect effects, one that is applicable in both linear and nonlinear models and that has broader policy-related interpretations. The paper establishes conditions under which such assessments can be estimated consistently from experimental and nonexperimental data, and thus extends path-analytic techniques to nonlinear and nonparametric models.

## 1 INTRODUCTION

The distinction between total, direct, and indirect effects is deeply entrenched in causal conversations, and attains practical importance in many applications, including policy decisions, legal definitions and health care analysis. Structural equation modeling (SEM) (Goldberger 1972), which provides a methodology of defining and estimating such effects, has been restricted to linear analysis, and no comparable methodology has been devised to extend these capabilities to models involving nonlinear dependencies,[1] as those commonly used in AI applications (Hagenaars 1993, p. 17).

The causal relationship that is easiest to interpret, define and estimate is the *total effect*. Written as $P(Y_x = y)$, the total effect measures the probability that response variable $Y$ would take on the value $y$ when $X$ is set to $x$ by external intervention.[2] This probability function is what we normally assess in a controlled experiment in which $X$ is randomized and in which the distribution of $Y$ is estimated for each level $x$ of $X$.

In many cases, however, this quantity does not adequately represent the target of investigation and attention is focused instead on the direct effect of $X$ on $Y$. The term "direct effect" is meant to quantify an influence that is not mediated by other variables in the model or, more accurately, the sensitivity of $Y$ to changes in $X$ while all other factors in the analysis are held fixed. Naturally, holding those factors fixed would sever all causal paths from $X$ to $Y$ with the exception of the direct link $X \to Y$, which is not intercepted by any intermediaries.

Indirect effects cannot be define in this manner, because it is impossible to hold a set of variables constant in such a way that the effect of $X$ on $Y$ measured under those conditions would circumvent the direct pathway, if such exists. Thus, the definition of indirect effects has remained incomplete, and, save for asserting inequality between direct and total effects, the very concept of "indirect effect" was deemed void of operational meaning (Pearl 2000, p. 165).

This paper shows that it is possible to give an operational meaning to both direct and indirect effects without fixing variables in the model, thus extending the applicability of these concepts to nonlinear and nonparametric models. The proposed generalization is based on a more subtle interpretation of "effects", here called "descriptive" (see Section 2.2), which concerns the action of causal forces under natural, rather than experimental conditions, and provides answers to a broader class of policy-related questions. This interpretation yields the standard path-coefficients in linear models, but leads to different formal definitions and different estimation procedures of direct and indirect effects in nonlinear models.

---

[1] A notable exception is the counterfactual analysis of Robins and Greenland (1992) which is applicable to non-linear models, but does not incorporate path-analytic techniques.

[2] The substripted notation $Y_x$ is borrowed from the potential-outcome framework of Rubin (1974). Pearl (2000) used, interchangeably, $P_x(y), P(y|do(x)), P(y|\hat{x})$, and $P(y_x)$, and showed their equivalence to probabilities of subjunctive conditionals: $P((X = x) \,\square\!\!\to (Y = y))$ (Lewis 1973).

Following a conceptual discussion of the descriptive and prescriptive interpretations (Section 2.2), Section 2.3 illustrates their distinct roles in decision-making contexts, while Section 2.4 discusses the descriptive basis and policy implications of indirect effects. Sections 3.2 and 3.3 provide, respectively, mathematical formulation of the prescriptive and descriptive interpretations of direct effects, while Section 3.4 establishes conditions under which the descriptive (or "natural") interpretation can be estimated consistently from either experimental or nonexperimental data. Sections 3.5 and 3.6 extend the formulation and identification analysis to indirect effects. In Section 3.7, we generalize the notion of indirect effect to *path-specific effects*, that is, effects transmitted through any specified set of paths in the model.

## 2 CONCEPTUAL ANALYSIS

### 2.1 Direct versus Total Effects

A classical example of the ubiquity of direct effects (Hesslow 1976) tells the story of a birth-control pill that is suspect of producing thrombosis in women and, at the same time, has a negative indirect effect on thrombosis by reducing the rate of pregnancies (pregnancy is known to encourage thrombosis). In this example, interest is focused on the direct effect of the pill because it represents a stable biological relationship that, unlike the total effect, is invariant to marital status and other factors that may affect women's chances of getting pregnant or of sustaining pregnancy. This invariance makes the direct effect transportable across cultural and sociological boundaries and, hence, a more useful quantity in scientific explanation and policy analysis.

Another class of examples involves legal disputes over race or sex discrimination in hiring. Here, neither the effect of sex or race on applicants' qualification nor the effect of qualification on hiring are targets of litigation. Rather, defendants must prove that sex and race do not *directly* influence hiring decisions, whatever indirect effects they might have on hiring by way of applicant qualification. This is made quite explicit in the following court ruling:

> "The central question in any employment-discrimination case is whether the employer would have taken the same action had the employee been of a different race (age, sex, religion, national origin etc.) and everything else had been the same." (Carson versus Bethlehem Steel Corp., 70 FEP Cases 921, 7th Cir. (1996), Quoted in Gastwirth 1997.)

Taking this criterion as a guideline, the direct effect of $X$ on $Y$ (in our case $X$=gender $Y$=hiring) can roughly be defined as the response of $Y$ to change in $X$ (say from $X = x^*$ to $X = x$) while keeping all other accessible variables at their initial value, namely, the value they would have attained under $X = x^*$.[3] This doubly-hypothetical criterion will be given precise mathematical formulation in Section 3, using the language and semantics of structural counterfactuals (Pearl 2000; chapter 7).

As a third example, one that illustrates the policy-making ramifications of direct and total effects, consider a drug treatment that has a side effect – headache. Patients who suffer from headache tend to take aspirin which, in turn may have its own effect on the disease or, may strengthen (or weaken) the impact of the drug on the disease. To determine how beneficial the drug is to the population as a whole, under existing patterns of aspirin usage, the total effect of the drug is the target of analysis, and the difference $P(Y_x = y) - P(Y_{x^*} = y)$ may serve to assist the decision, with $x$ and $x^*$ being any two treatment levels. However, to decide whether aspirin should be encouraged or discouraged during the treatment, the *direct* effect of the drug on the disease, both with aspirin and without aspirin, should be the target of investigation. The appropriate expression for analysis would then be the difference $P(Y_{xz} = y) - P(Y_{x^*z} = y)$, where $z$ stands for any specified level of aspirin intake.

In linear systems, direct effects are fully specified by the corresponding path coefficients, and are independent of the values at which we hold the the intermediate variables ($Z$ in our examples). In nonlinear systems, those values would, in general, modify the effect of $X$ on $Y$ and thus should be chosen carefully to represent the target policy under analysis. This lead to a basic distinction between two types of conceptualizations: *prescriptive* and *descriptive*.

---

[3] Robins and Greenland (1992) have adapted essentially the same criterion (phrased differently) for their interpretation of "direct effect" in epidemiology.

## 2.2 Descriptive versus prescriptive interpretation

We will illustrate this distinction using the treatment-aspirin example described in the last section. In the prescriptive conceptualization, we ask whether a specific untreated patient would improve if treated, while holding the aspirin intake fixed at some predetermined level, say $Z = z$. In the descriptive conceptualization, we ask again whether the untreated patient would improve if treated, but now we hold the aspirin intake fixed at whatever level the patient currently consumes under no-treatment condition. The difference between these two conceptualizations lies in whether we wish to account for the natural relationship between the direct and the mediating cause (that is, between treatment and aspirin) or to modify that relationship to match policy objectives. We call the effect computed from the descriptive perspective the *natural* effect, and the one computed from the prescriptive perspective the *controlled* effect.

Consider a patient who takes aspirin if and only if treated, and for whom the treatment is effective only when aspirin is present. For such a person, the treatment is deemed to have no natural direct effect (on recovery), because, by keeping the aspirin at the current, pre-treatment level of zero, we ensure that the treatment effect would be nullified. The controlled direct effect, however, is nonzero for this person, because the efficacy of the treatment would surface when we fix the aspirin intake at non-zero level. Note that the descriptive formulation requires knowledge of the individual natural behavior—in our example, whether the untreated patient actually uses aspirin—while the prescriptive formulation requires no such knowledge.

This difference becomes a major stumbling block when it comes to estimating *average* direct effects in a population of individuals. At the population level, the prescriptive formulation is pragmatic; we wish to predict the difference in recovery rates between treated and untreated patients when a prescribed dose of aspirin is administered to all patients in the population—the actual consumption of aspirin under uncontrolled conditions need not concern us. In contrast, the descriptive formulation is attributional; we ask whether an observed improvement in recovery rates (again, between treated and untreated patients) is attributable to the treatment itself, as opposed to preferential use of aspirin among treated patients. To properly distinguish between these two contributions, we therefore need to measure the improvement in recovery rates while making each patient take the same level of aspirin that he/she took before treatment. However, as Robins and Greenland (1992) pointed out, such control over individual behavior would require testing the same group of patients twice (i.e., under treatment and no treatment conditions), and cannot be administered in experiments with two different groups, however randomized. (There is no way to determine what level of aspirin an untreated patient would take if treated, unless we actually treat that patient and, then, this patient could no longer be eligible for the untreated group.) Since repeatable tests on the same individuals are rarely feasible, the descriptive measure of the direct effect is not generally estimable from standard experimental studies. In Section 3.4 we will analyze what additional assumptions are required for consistently estimating this measure, the *average natural direct effect*, from either experimental or observational studies.

## 2.3 Policy implications of the Descriptive interpretation

Why would anyone be interested in assessing the average natural direct effect? Assume that the drug manufacturer is considering ways of eliminating the adverse side-effect of the drug, in our case, the headache. A natural question to ask is whether the drug would still retain its effectiveness in the population of interest. The controlled direct effect would not give us the answer to this question, because it refers to a specific aspirin level, taken uniformly by all individuals. Our target population is one where aspirin intake varies from individual to individual, depending on other factors beside drug-induced headache, factors which may also cause the effectiveness of the drug to vary from individual to individual. Therefore, the parameter we need to assess is the average natural direct effect, as described in the Subsection 2.2.

This example demonstrates that the descriptive interpretation of direct effects is not purely "descriptive"; it carries a definite operational implications, and answers policy-related questions of practical significance. Moreover, note that the policy question considered in this example cannot be represented in the standard syntax of $do(x)$ operators—it does not involve fixing any of the variables in the model but, rather, modifying the causal paths in the model. Even if "headache" were a genuine variable in our model, the elimination of any drug-induced headache is not equivalent to setting any "headache" to zero, since a person might get headache for reason other than the drug. Instead, the policy option involves the de-activation of the causal path from "drug" to "headache".

In general, the average natural direct effect would be of interest in evaluating policy options of a more refined variety, ones that involve, not merely fixing the levels of the variables in the model, but also determining how these levels would influence one another.

Typical examples of such options involve choosing the *manner* (e.g., instrument, or timing) in which a given decision is implemented, or choosing the agents that should be *informed* about the decision. A firm often needs to assess, for example, whether it would be worthwhile to conceal a certain decision from a competitor. This amounts, again, to evaluating the natural direct effect of the decision in question, unmediated by the competitor's reaction. Theoretically, such policy options could conceivably be represented as (values of) variables in a more refined model, for example one where the concept "the effect of treatment on headache" would be given a variable name, and where the manufacturer decision to eliminate side-effects would be represented by fixing this hypothetical variable to zero. The analysis of this paper shows that such unnatural modeling techniques can be avoided, and that important nonstandard policy questions can be handled by standard models, where variables stands for directly measurable quantities.

### 2.4 Descriptive interpretation of indirect effects

The descriptive conception of direct effects can easily be transported to the formulation of indirect effects; oddly, the prescriptive formulation is not transportable. Returning to our treatment-aspirin example, if we wish to assess the *natural* indirect effect of treatment on recovery for a specific patient, we withhold treatment and ask, instead, whether that patient would recover if given as much aspirin as he/she would have taken if he/she had been under treatment. In this way, we insure that whatever changes occur in the patient's condition are due to treatment-induced aspirin consumption and not to the treatment itself. Similarly, at the population level, the natural indirect effect of the treatment is interpreted as the improvement in recovery rates if we were to withhold treatment from all patients but, instead, let each patient take the same level of aspirin that he/she would have taken under treatment. As in the descriptive formulation of direct effects, this hypothetical quantity involves nested counterfactuals and will be identifiable only under special circumstances.

The prescriptive formulation has no parallel in indirect effects, for reasons discussed in the introduction section; there is no way of preventing the direct effect from operating by holding certain variables constant. We will see that, in linear systems, the descriptive and prescriptive formulations of direct effects lead, indeed, to the same expression in terms of path coefficients. The corresponding linear expression for indirect effects, computed as the difference between the total and direct effects, coincides with the descriptive formulation but finds no prescriptive interpretation.

The operational implications of indirect effects, like those of natural direct effect, concern nonstandard policy options. Although it is impossible, by controlling variables, to block a direct path (i.e., a single edge), if such exists, it is nevertheless possible to block such a path by more refined policy options, ones that deactivate the direct path through the manner in which an action is taken or through the mode by which a variable level is achieved. In the hiring discrimination example, if we make it illegal to question applicants about their gender, (and if no other indication of gender are available to the hiring agent), then any residual sex preferences (in hiring) would be attributable to the indirect effect of sex on hiring. A policy maker might well be interested in predicting the magnitude of such preferences from data obtained prior to implementing the no-questioning policy, and the average indirect effect would then provide the sought for prediction. A similar refinement applies in the firm-competitor example of the preceding subsection. A firm might wish to assess, for example, the economical impact of bluffing a competitor into believing that a certain decision has been taken by the firm, and this could be implemented by (secretly) instructing certain agents to ignore the decision. In both cases, our model may not be sufficiently detailed to represents such policy options in the form of variable fixing (e.g., the agents may not be represented as intermediate nodes between the decision and its effect) and the task amounts then to evaluating the average natural indirect effects in a coarse-grain model, where a direct link exists between the decision and its outcome.

## 3 FORMAL ANALYSIS

### 3.1 Notation

Throughout our analysis we will let $X$ be the control variable (whose effect we seek to assess), and let $Y$ be the response variable. We will let $Z$ stand for the set of all intermediate variables between $X$ and $Y$ which, in the simplest case considered, would be a single variable as in Figure 1(a). Most of our results will still be valid if we let $Z$ stand for any set of such variables, in particular, the set of $Y$'s parents excluding $X$.

We will use the counterfactual notation $Y_x(u)$ to denote the value that $Y$ would attain in unit (or situation) $U = u$ under the control regime $do(X = x)$. See Pearl (2000, Chapter 7) for formal semantics of these counterfactual utterances. Many concepts associated with direct and indirect effect require comparison to a reference value of $X$, that is, a value relative to which we measure changes. We will designate this reference value by $x^*$.

### 3.2 Controlled Direct Effects (review)

**Definition 1** (*Controlled unit-level direct-effect; qualitative*)

A variable $X$ is said to have a controlled direct effect on variable $Y$ in model $M$ and situation $U = u$ if there exists a setting $Z = z$ of the other variables in the model and two values of $X$, $x^*$ and $x$, such that

$$Y_{x^*z}(u) \neq Y_{xz}(u) \qquad (1)$$

In words, the value of $Y$ under $X = x^*$ differs from its value under $X = x$ when we keep all other variables $Z$ fixed at $z$. If condition (1) is satisfied for some $z$, we say that the transition event $X = x$ has a controlled direct-effect on $Y$, keeping the reference point $X = x^*$ implicit.

Clearly, confining $Z$ to the parents of $Y$ (excluding $X$) leaves the definition unaltered.

**Definition 2** (*Controlled unit-level direct-effect; quantitative*)
Given a causal model $M$ with causal graph $G$, the controlled direct effect of $X = x$ on $Y$ in unit $U = u$ and setting $Z = z$ is given by

$$CDE_z(x, x^*; Y, u) = Y_{xz}(u) - Y_{x^*z}(u) \qquad (2)$$

where $Z$ stands for all parents of $Y$ (in $G$) excluding $X$.

Alternatively, the ratio $Y_{xz}(u)/Y_{x^*z}(u)$, the proportional difference $(Y_{xz}(u) - Y_{x^*z}(u))/Y_{x^*z}(u)$, or some other suitable relationship might be used to quantify the magnitude of the direct effect; the difference is by far the most common measure, and will be used throughout this paper.

**Definition 3** (*Average controlled direct effect*)
Given a probabilistic causal model $\langle M, P(u) \rangle$, the controlled direct effect of event $X = x$ on $Y$ is defined as:

$$CDE_z(x, x^*; Y) = E(Y_{xz} - Y_{x^*z}) \qquad (3)$$

where the expectation is taken over $u$.

The distribution $P(Y_{xz} = y)$ can be estimated consistently from experimental studies in which both $X$ and $Z$ are randomized. In nonexperimental studies, the identification of this distribution requires that certain "no-confounding" assumptions hold true in the population tested. Graphical criteria encapsulating these assumptions are described in Pearl (2000, Sections 4.3 and 4.4).

### 3.3 Natural Direct Effects: Formulation

**Definition 4** (*Unit-level natural direct effect; qualitative*)
An event $X = x$ is said to have a natural direct effect on variable $Y$ in situation $U = u$ if the following inequality holds

$$Y_{x^*}(u) \neq Y_{x, Z_{x^*}(u)}(u) \qquad (4)$$

In words, the value of $Y$ under $X = x^*$ differs from its value under $X = x$ even when we keep $Z$ at the same value $(Z_{x^*}(u))$ that $Z$ attains under $X = x^*$.

We can easily extend this definition from events to variables by defining $X$ as having a *natural* direct effect on $Y$ (in model $M$ and situation $U = u$) if there exist two values, $x^*$ and $x$, that satisfy (4). Note that this definition no longer requires that we specify a value $z$ for $Z$; that value is determined naturally by the model, once we specify $x, x^*$, and $u$. Note also that condition (4) is a direct literal translation of the court criterion of sex discrimination in hiring (Section 2.1) with $X = x^*$ being a male, $X = x$ a female, $Y = 1$ a decision to hire, and $Z$ the set of all other attributes of individual $u$.

If one is interested in the magnitude of the natural direct effect, one can take the difference

$$Y_{x, Z_{x^*}(u)}(u) - Y_{x^*}(u) \qquad (5)$$

and designate it by the symbol $NDE(x, x^*; Y, u)$ (acronym for Natural Direct Effect). If we are further interested in assessing the average of this difference in a population of units, we have:

**Definition 5** (*Average natural direct effect*)
The average natural direct effect of event $X = x$ on a response variable $Y$, denoted $NDE(x, x^*; Y)$, is defined as

$$NDE(x, x^*; Y) = E(Y_{x, Z_{x^*}}) - E(Y_{x^*}) \qquad (6)$$

Applied to the sex discrimination example of Section 2.1, (with $x^* =$ male, $x =$ female, $y =$ hiring, $z =$ qualifications) Eq. (6) measures the expected change in male hiring, $E(Y_{x^*})$, if employers were instructed to treat males' applications as though they were females'.

### 3.4 Natural Direct Effects: Identification

As noted in Section 2, we cannot generally evaluate the average natural direct-effect from empirical data. Formally, this means that Eq. (6) is not reducible to expressions of the form

$$P(Y_x = y) \text{ or } P(Y_{xz} = y);$$

the former governs the causal effect of $X$ on $Y$ (obtained by randomizing $X$) and the latter governs the causal effect of $X$ and $Z$ on $Y$ (obtained by randomizing both $X$ and $Z$).

We now present conditions under which such reduction is nevertheless feasible.

**Theorem 1** (*Experimental identification*)
*If there exists a set $W$ of covariates, nondescendants of $X$ or $Z$, such that*

$$Y_{xz} \perp\!\!\!\perp Z_{x^*} | W \qquad \text{for all } z \text{ and } x \qquad (7)$$

(*read: $Y_{xz}$ is conditionally independent of $Z_{x^*}$, given $W$*), *then the average natural direct-effect is experimentally identifiable, and it is given by*

$$NDE(x, x^*; Y) = \sum_{w,z} [E(Y_{xz}|w) - E(Y_{x^*z}|w)] P(Z_{x^*} = z|w) P(w) \qquad (8)$$

**Proof**
The first term in (6) can be written

$$E(Y_{x,Z_{x^*}}) = \sum_w \sum_z E(Y_{xz}|Z_{x^*} = z, W = w) P(Z_{x^*} = z|W = w) P(W = w) \qquad (9)$$

Using (7), we obtain:

$$E(Y_{x,Z_{x^*}}) = \sum_w \sum_z E(Y_{xz} = y|W = w) P(Z_{x^*} = z|W = w) P(W = w) \qquad (10)$$

Each factor in (10) is identifiable; $E(Y_{xz} = y|W = w)$, by randomizing $X$ and $Z$ for each value of $W$, and $P(Z_{x^*} = z|W = w)$ by randomizing $X$ for each value of $W$. This proves the assertion in the theorem. Substituting (10) into (6) and using the law of composition $E(Y_{x^*}) = E(Y_{x^* Z_{x^*}})$ (Pearl 2000, p. 229) gives (8), and completes the proof of Theorem 1. □

The conditional independence relation in Eq. (7) can easily be verified from the causal graph associated with the model. Using a graphical interpretation of counterfactuals (Pearl 2000, p. 214-5), this relation reads:

$$(Y \perp\!\!\!\perp Z | W)_{G_{\underline{XZ}}} \qquad (11)$$

In words, $W$ $d$-separates $Y$ from $Z$ in the graph formed by deleting all (solid) arrows emanating from $X$ and $Z$.

Figure 1(a) illustrates a typical graph associated with estimating the direct effect of $X$ on $Y$. The identifying subgraph is shown in Fig. 1(b), and illustrates how $W$ $d$-separates $Y$ from $Z$. The separation condition in (11) is somewhat stronger than (7), since the former implies the latter for every pair of values, $x$ and $x^*$, of $X$ (see (Pearl 2000, p. 214)). Likewise, condition (7) can be relaxed in several ways. However, since assumptions of counterfactual independencies can be meaningfully substantiated only when cast in structural form (Pearl 2000, p. 244–5), graphical conditions will be the target of our analysis.

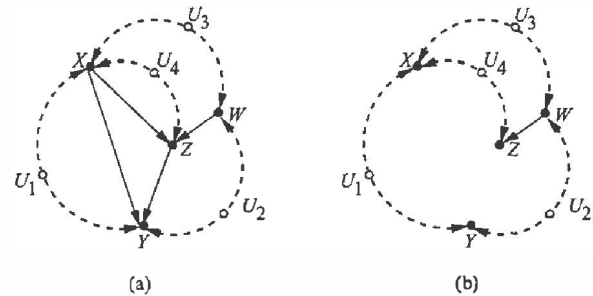

Figure 1: (a) A causal model with latent variables ($U$'s) where the natural direct effect can be identified in experimental studies. (b) The subgraph $G_{\underline{XZ}}$ illustrating the criterion of experimental identifiability (Eq. 11): $W$ $d$-separates $Y$ from $Z$.

The identification of the natural direct effect from *nonexperimental* data requires stronger conditions. From Eq. (8) we see that it is sufficient to identify the conditional probabilities of two counterfactuals: $P(Y_{xz} = y|W = w)$ and $P(Z_{x^*} = z|W = w)$, where $W$ is any set of covariates that satisfies Eq. (7) (or (11)). This yields the following criterion for identification:

**Theorem 2** (*Nonexperimental identification*)
*The average natural direct-effect $NDE(x, x^*; Y)$ is identifiable in nonexperimental studies if there exists a set $W$ of covariates, nondescendants of $X$ or $Z$, such that, for all values $z$ and $x$ we have:*

(*i*) $Y_{xz} \perp\!\!\!\perp Z_{x^*} | W$

(*ii*) $P(Y_{xz} = y|W = w)$ *is identifiable*

(*iii*) $P(Z_{x^*} = z|W = w)$ *is identifiable*

*Moreover, if conditions (i)-(iii) are satisfied, the natural direct effect is given by (8).*

Explicating these identification conditions in graphical terms (using Theorem 4.41 in (Pearl 2000)) yields the following corollary:

**Corollary 1** (*Graphical identification criterion*)
The average natural direct-effect $NDE(x, x^*; Y)$ is identifiable in nonexperimental studies if there exist four sets of covariates, $W_0, W_1, W_2,$ and $W_3$, such that

(i) $(Y \perp\!\!\!\perp Z | W_0)_{G_{\underline{XZ}}}$

(ii) $(Y \perp\!\!\!\perp X | W_0, W_1)_{G_{\underline{X}\overline{Z}}}$

(iii) $(Y \perp\!\!\!\perp Z | X, W_0, W_1, W_2)_{G_{\underline{Z}}}$

(iv) $(Z \perp\!\!\!\perp X | W_0, W_3)_{G_{\underline{X}}}$

(v) $W_0, W_1,$ and $W_3$ contain no descendant of $X$ and $W_2$ contains no descendant of $Z$.

(Remark: $G_{\underline{X}\overline{Z}}$ denotes the graph formed by deleting (from $G$) all arrows emanating from $X$ or entering $Z$.)

As an example for applying these criteria, consider Figure 1(a), and assume that all variables (including the $U$'s) are observable. Conditions (i)-(iv) of Corollary 1 are satisfied if we choose:

$W_0 = \{W\}, W_1 = \{U_1, U_2\}, W_2 = \emptyset$ and $W_3 = \{U_4\}$

or, alternatively,

$W_0 = \{U_2\}, W_1 = \{U_1\}, W_2 = \emptyset$ and $W_3 = \{U_3, U_4\}$

It is instructive to examine the form that expression (8) takes in Markovian models, (that is, acyclic models with independent error terms) where condition (7) is always satisfied with $W = \emptyset$, since $Y_{xz}$ is independent of all variables in the model. In Markovian models, we also have the following three relationships:

$$P(Y_{xz} = y) = P(y|x, z) \quad (12)$$

since $X \cup Z$ is the set of $Y$'s parents,

$$P(Z_{x^*} = z) = \sum_s P(z|x^*, s) P(s), \quad (13)$$

$$P(Y_{x, Z_{x^*}} = y) = \sum_s \sum_z P(y|x, z) P(z|x^*, s) P(s) \quad (14)$$

where $S$ stands for the parents of $Z$, excluding $X$, or any other set satisfying the back-door criterion (Pearl 2000, p. 79). This yields the following corollary of Theorem 1:

**Corollary 2** *The average natural direct effect in Markovian models is identifiable from nonexperimental data, and it is given by*

$$NDE(x, x^*; Y) = \sum_s \sum_z [E(Y|x, z) - E(Y|x^*, z)] P(z|x^*, s) P(s) \quad (15)$$

where $S$ stands for any set satisfying the back-door criterion between $X$ and $Z$.

Eq. (15) follows by substituting (14) into (6) and using the identity $E(Y_{x^*}) = E(Y_{x^*, Z_{x^*}})$.

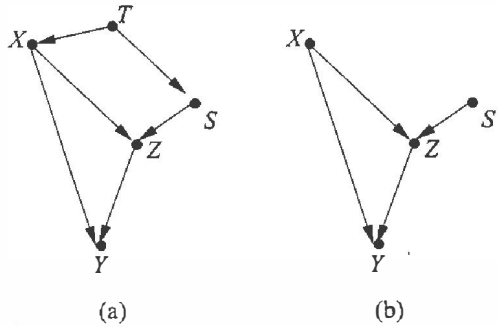

(a) (b)

Figure 2: Simple Markovian models for which the natural direct effect is given by Eq. (15) (for (a)) and Eq. (17) (for (b)).

Further insight can be gained by examining simple Markovian models in which the effect of $X$ on $Z$ is not confounded, that is,

$$P(Z_{x^*} = z) = P(z|x^*) \quad (16)$$

In such models, a simple version of which is illustrated in Fig. 2(b), Eq. (13) can be replace by (16) and (15) simplifies to

$$NDE(x, x^*; Y) = \sum_z [E(Y|x, z) - E(Y|x^*, z)] P(z|x^*) \quad (17)$$

This expression has a simple interpretation as a weighted average of the controlled direct effect $E(Y|x, z) - E(Y|x^*, z)$, where the intermediate value $z$ is chosen according to its distribution under $x^*$.

### 3.5 Natural Indirect Effects: Formulation

As we discussed in Section 2.4, the prescriptive formulation of "controlled direct effect" has no parallel in indirect effects; we therefore use the descriptive formulation, and define *natural* indirect effects at both the unit and population levels. Lacking the controlled alternative, we will drop the title "natural" from discussions of indirect effects, unless it serves to convey a contrast.

**Definition 6** (*Unit-level indirect effect; qualitative*)
An event $X = x$ is said to have an indirect effect on variable $Y$ in situation $U = u$ if the following inequality holds

$$Y_{x^*}(u) \neq Y_{x^*, Z_x(u)}(u) \quad (18)$$

*In words, the value of $Y$ changes when we keep $X$ fixed at its reference level $X = x^*$ and change $Z$ to a new value, $Z_x(u)$, the same value that $Z$ would attain under $X = x$.*

Taking the difference between the two sides of Eq. (18), we can define the unit level indirect effect as

$$NIE(x, x^*; Y, u) = Y_{x^*, Z_x(u)}(u) - Y_{x^*}(u) \quad (19)$$

and proceed to define its average in the population:

**Definition 7** (*Average indirect effect*)
*The average indirect effect of event $X = x$ on variable $Y$, denoted $NIE(x, x^*; Y)$, is defined as*

$$NIE(x, x^*; Y) = E(Y_{x^*, Z_x}) - E(Y_{x^*}) \quad (20)$$

Comparing Eqs. (6) and (20), we see that the indirect effect associated with the transition from $x^*$ to $x$ is closely related to the natural direct effect associated with the reverse transition, from $x$ to $x^*$. In fact, recalling that the difference $E(Y_x) - E(Y_{x^*})$ equals the total effect of $X = x$ on $Y$,

$$TE(x, x^*; Y) = E(Y_x) - E(Y_{x^*}) \quad (21)$$

we obtain the following theorem:

**Theorem 3** *The total, direct and indirect effects obey the following relationships*

$$TE(x, x^*; Y) = NIE(x, x^*; Y) - NDE(x^*, x; Y) \quad (22)$$
$$TE(x, x^*; Y) = NDE(x, x^*; Y) - NIE(x^*, x; Y) \quad (23)$$

*In words, the total effect (on $Y$) associated with the transition from $x^*$ to $x$ is equal to the difference between the indirect effect associated with this transition and the (natural) direct effect associated with the reverse transition, from $x$ to $x^*$.*

As strange as these relationships appear, they produce the standard, additive relation

$$TE(x, x^*; Y) = NIE(x, x^*; Y) + NDE(x, x^*; Y) \quad (24)$$

when applied to linear models. The reason is clear; in linear systems the effect of the transition from $x^*$ to $x$ is proportional to $x - x^*$, hence it is always equal and of opposite sign to the effect of the reverse transition. Thus, substituting in (22) (or (23)), yields (24).

### 3.6 Natural Indirect Effects: Identification

Eqs. (22) and (23) show that the indirect effect is identified whenever both the total and the (natural) direct effect are identified (for all $x$ and $x^*$). Moreover, the identification conditions and the resulting expressions for indirect effects are identical to the corresponding ones for direct effects (Theorems 1 and 2), save for a simple exchange of the indices $x$ and $x^*$. This is explicated in the following theorem.

**Theorem 4** *If there exists a set $W$ of covariates, nondescendants of $X$ or $Z$, such that*

$$Y_{x^*z} \perp\!\!\!\perp Z_x | W \quad (25)$$

*for all $x$ and $z$, then the average indirect-effect is experimentally identifiable, and it is given by*

$$NIE(x, x^*; Y)$$
$$= \sum_{w,z} E(Y_{x^*z}|w)[P(Z_x = z|w) - P(Z_{x^*} = z|w)]P(w)$$
$$(26)$$

*Moreover, the average indirect effect is identified in nonexperimental studies whenever the following expressions are identified for all $z$ and $w$:*

$$E(Y_{x^*z}|w), \ P(Z_x = z|w) \text{ and } P(Z_{x^*} = z|w),$$

*with $W$ satisfying Eq. (25).*

In the simple Markovian model depicted in Fig. 2(b), Eq. (26) reduces to

$$NIE(x, x^*; Y) = \sum_z E(Y|x^*, z)[P(z|x) - P(z|x^*)] \quad (27)$$

Contrasting Eq. (27) with Eq. (17), we see that the expression for the indirect effect fixes $X$ at the reference value $x^*$, and lets $z$ vary according to its distribution under the post-transition value of $X = x$. The expression for the direct effect fixes $X$ at $x$, and lets $z$ vary according to its distribution under the reference conditions $X = x^*$.

Applied to the sex discrimination example of Section 2.1, Eq. (27) measures the expected change in male hiring, $E(Y_{x^*})$, if males were trained to acquire (in distribution) equal qualifications ($Z = z$) as those of females ($X = x$).

### 3.7 General Path-specific Effects

The analysis of the last section suggests that path-specific effects can best be understood in terms of a *path-deactivation process*, where a selected set of paths, rather than nodes, are forced to remain inactive during the transition from $X = x^*$ to $X = x$. In Figure 3, for example, if we wish to evaluate the effect of $X$ on $Y$ transmitted by the subgraph $g : X \to Z \to W \to Y$, we cannot hold $Z$ or $W$ constant, for both must vary in the process. Rather, we isolate the desired effect by fixing the appropriate subset of arguments in each equation. In other words, we replace $x$ with $x^*$ in the equation for $W$, and replace $z$ with $z^*(u) = Z_{x^*}(u)$ in the equation for $Y$. This amounts to creating a new model, in which each structural function $f_i$ in $M$ is replaced with a new function of a smaller set of arguments, since some of the arguments are replaced by constants. The following definition expresses this idea formally.

**Definition 8** (*path-specific effect*)
*Let $G$ be the the causal graph associated with model $M$, and let $g$ be an edge-subgraph of $G$ containing the paths selected for effect analysis. The $g$-specific effect of $x$ on $Y$ (relative to reference $x^*$) is defined as the total effect of $x$ on $Y$ in a modified model $M_g^*$ formed as follows. Let each parent set $PA_i$ in $G$ be partitioned into two parts*

$$PA_i = \{PA_i(g), PA_i(\overline{g})\} \qquad (28)$$

*where $PA_i(g)$ represents those members of $PA_i$ that are linked to $X_i$ in $g$, and $PA_i(\overline{g})$ represents the complementary set, from which there is no link to $X_i$ in $g$. We replace each function $f_i(pa_i, u)$ with a new function $f_i^*(pa_i, u; g)$, defined as*

$$f_i^*(pa_i, u; g) = f_i(pa_i(g), pa_i^*(\overline{g}), u) \qquad (29)$$

*where $pa_i^*(\overline{g})$ stands for the values that the variables in $PA_i(\overline{g})$ would attain (in $M$ and $u$) under $X = x^*$ (that is, $pa_i^*(\overline{g}) = PA_i(\overline{g})_{x^*}$). The $g$-specific effect of $x$ on $Y$, denoted $SE_g(x, x^*; Y, u)_M$ is defined as*

$$SE_g(x, x^*; Y, u)_M = TE(x, x^*; Y, u)_{M_g^*}. \qquad (30)$$

We demonstrate this construction in the model of Fig. 3 which stands for the equations:

$$\begin{aligned} z &= f_Z(x, u_Z) \\ w &= f_W(z, x, u_W) \\ y &= f_Y(z, w, u_Y) \end{aligned}$$

where $u_Z, u_W$, and $u_Y$ are the components of $u$ that enter the corresponding equations. Defining $z^*(u) = f_Z(x^*, u_Z)$, the modified model $M_g^*$ reads:

$$\begin{aligned} z &= f_Z(x, u_Z) \\ w &= f_W(z, x^*, u_W) \text{ and} \\ y &= f_Y(z^*(u), w, u_Y) \end{aligned} \qquad (31)$$

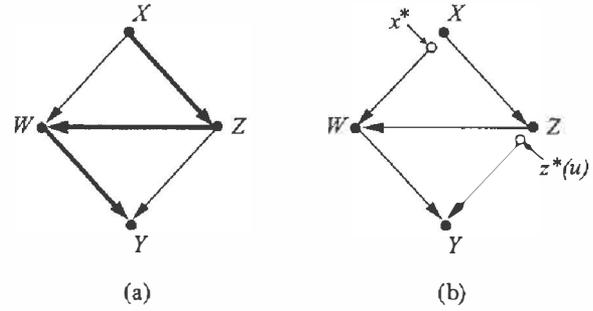

Figure 3: The path-specific effect transmitted through $X \to Z \to W \to Y$ (heavy lines) in (a) is equal to the total effect transmitted through the model in (b), treating $x^*$ and $z^*(u)$ as constants. (By convention, $u$ is not shown in the diagram.)

and our task amounts to computing the total effect of $x$ on $Y$ in $M_g^*$, or

$$\begin{aligned} TE(x, x^*; Y, u)_{M_g^*} &= \\ &= f_Y(z^*(u), f_W(f_Z(x, u_Z), x^*, u_W), u_Y) \\ &\quad - Y_{x^*}(u) \end{aligned} \qquad (32)$$

It can be shown that the identification conditions for general path-specific effects are much more stringent than those of the direct and indirect effects. The path-specific effect shown in Figure 3, for example, is not identified even in Markovian models. Since direct and indirect effects are special cases of path-specific effects, the identification conditions of Theorems 2 and 3 raise the interesting question of whether a simple characterization exists of the class of subgraphs, $g$, whose path-specific effects are identifiable in Markovian models. I hope inquisitive readers will be able to solve this open problem.

## 4 Conclusions

This paper formulates a new definition of path-specific effects that is based on path switching, instead of variable fixing, and that extends the interpretation and evaluation of direct and indirect effects to nonlinear models. It is shown that, in nonparametric models, direct and indirect effects can be estimated consistently from both experimental and nonexperimental data, provided certain conditions hold in the causal diagram. Markovian models always satisfy these conditions. Using the new definition, the paper provides an operational interpretation of indirect effects, the policy significance of which was deemed enigmatic in recent literature.

On the conceptual front, the paper uncovers a class of nonstandard policy questions that cannot be for-mulated in the usual variable-fixing vocabulary and that can be evaluated, nevertheless, using the notions of direct and indirect effects. These policy questions concern redirecting the flow of influence in the system, and generally involve the deactivation of existing influences among specific variables. The ubiquity and manageabiligy of such questions in causal modeling suggest that value-assignment manipulations, which control the outputs of the causal mechanism in the model, are less fundamental to the notion of causation than input-selection manipulations, which control the signals driving those mechanisms.

## Acknowledgment

My interest in this topic was stimulated by Jacques Hagenaars, who pointed out the importance of quantifying indirect effects in the social sciences (See ⟨http://bayes.cs.ucla.edu/BOOK-2K/hagenaars.html⟩.) Sol Kaufman, Sander Greenland, Steven Fienberg and Chris Hitchcock have provided helpful comments on the first draft of this paper. This research was supported in parts by grants from NSF, ONR (MURI) and AFOSR.